\documentclass{article}

\usepackage[preprint, nonatbib]{templates/NeurIPS/neurips_2022}
\usepackage[utf8]{inputenc} % allow utf-8 input
\usepackage[T1]{fontenc}    % use 8-bit T1 fonts
\usepackage{hyperref}       % hyperlinks
\usepackage{url}            % simple URL typesetting
\usepackage{booktabs}       % professional-quality tables
\usepackage{amsfonts}       % blackboard math symbols
\usepackage{nicefrac}       % compact symbols for 1/2, etc.
\usepackage{microtype}      % microtypography
\usepackage[table]{xcolor}         % colors
\usepackage{graphicx}
\usepackage{amsmath}
\usepackage{amssymb}
\usepackage{enumitem}
\usepackage{xfrac}
\usepackage[export]{adjustbox}
\usepackage{svg}
\usepackage{style}
\usepackage{pifont}
\usepackage{multirow}
\usepackage{mwe}
\usepackage{transparent}
\hypersetup{colorlinks=true}
\usepackage[capitalize]{cleveref}

\title{Detecting Looted Archaeological Sites\\ from Satellite Image Time Series}

\author{
  Elliot Vincent\textsuperscript{1, 2},~~~Mehraïl Saroufim\textsuperscript{3},~~~Jonathan Chemla\textsuperscript{3},~~~Yves Ubelmann\textsuperscript{3},\\\textbf{Philippe Marquis\textsuperscript{4},~~~Jean Ponce\textsuperscript{5, 6},~~~Mathieu Aubry\textsuperscript{1}}\\
  \textsuperscript{1}LIGM, Ecole des Ponts, Univ Gustave Eiffel, CNRS, France~~~\textsuperscript{2}Inria Paris\\
  \textsuperscript{3}Iconem~~~\textsuperscript{4}DAFA, French archaeological delegation in Afghanistan\\
  \textsuperscript{4}Department of Computer Science, Ecole normale supérieure (ENS-PSL, CNRS, Inria)\\
  \textsuperscript{5}Courant Institute of Mathematical Sciences and Center for Data Science, New York University\\
  \texttt{\{elliot.vincent,mathieu.aubry\}@enpc.fr},~~~\texttt{jean.ponce@ens.fr}\\\texttt{\{yu,jchemla,mehrail.saroufim\}@iconem.com}, ~~~\texttt{marquis.dafa@hotmail.fr}
  }

\begin{document}

\maketitle

\begin{abstract}
Archaeological sites are the physical remains of past human activity and one of the main sources of information about past societies and cultures. However, they are also the target of malevolent human actions, especially in countries having experienced inner turmoil and conflicts. Because monitoring these sites from space is a key step towards their preservation, we introduce the DAFA Looted Sites dataset, \datasetname, a labeled multi-temporal remote sensing dataset containing 55,480 images acquired monthly over 8 years across 675 Afghan archaeological sites, including 135 sites looted during the acquisition period. \datasetname~is particularly challenging because of the limited number of training samples, the class imbalance, the weak binary annotations only available at the level of the time series, and the subtlety of relevant changes coupled with important irrelevant ones over a long time period. It is also an interesting playground to assess the performance of satellite image time series (SITS) classification methods on a real and important use case. We evaluate a large set of baselines, outline the substantial benefits of using foundation models and show the additional boost that can be provided by using complete time series instead of using single images. The code and dataset are available at \url{https://github.com/ElliotVincent/DAFA-LS}.
\end{abstract}
\begin{figure}[t!]
    \renewcommand{\arraystretch}{1}
    \setlength\tabcolsep{2pt}
    \centering
        \resizebox{\linewidth}{!}{
     \begin{tabular}{ccc}
         \includegraphics[width=0.43\linewidth]{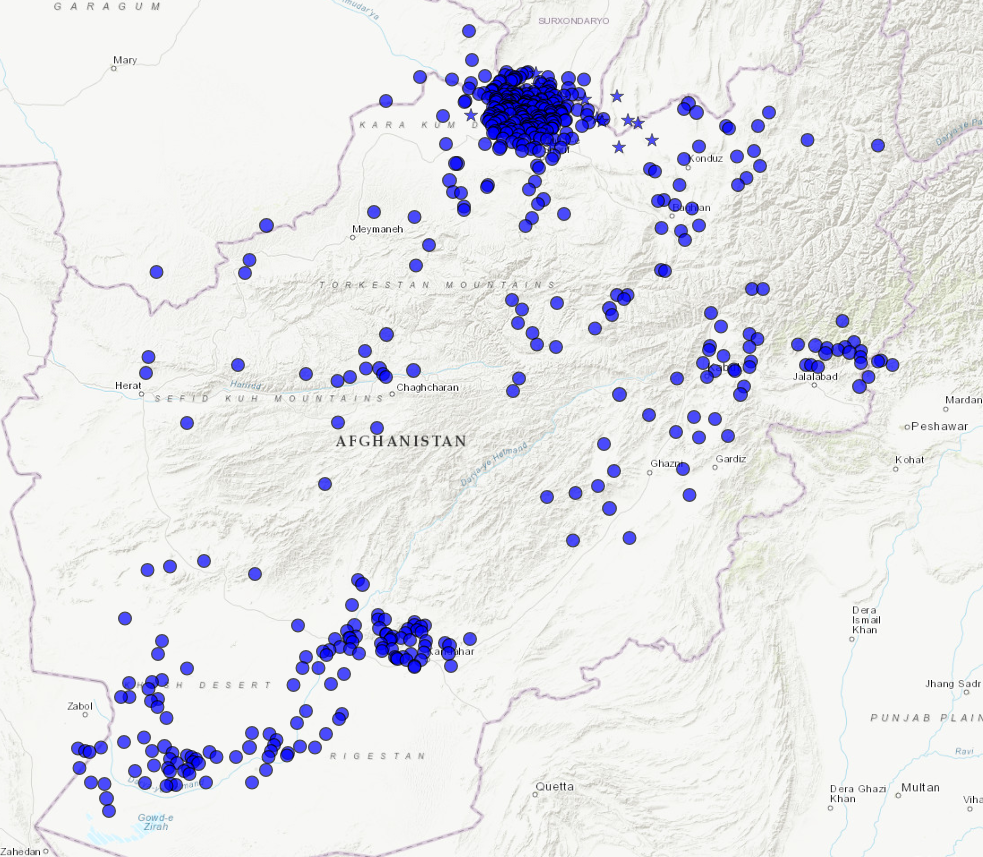} & & \includegraphics[width=0.43\linewidth]{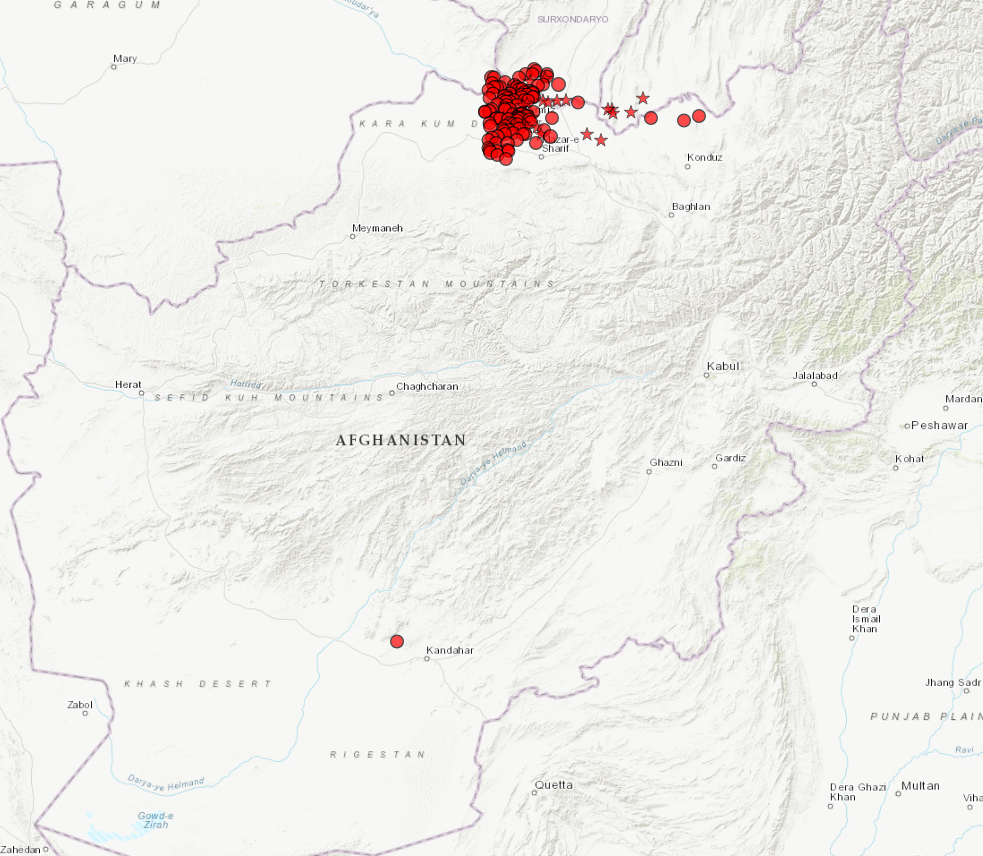}\\
         (a) Map of preserved sites & & (b) Map of looted sites
    \end{tabular}
    }
    \resizebox{\linewidth}{!}{
    \begin{tabular}{cccc}
         \multirow{2}{*}[20.5mm]{\includegraphics[width=0.43\linewidth]{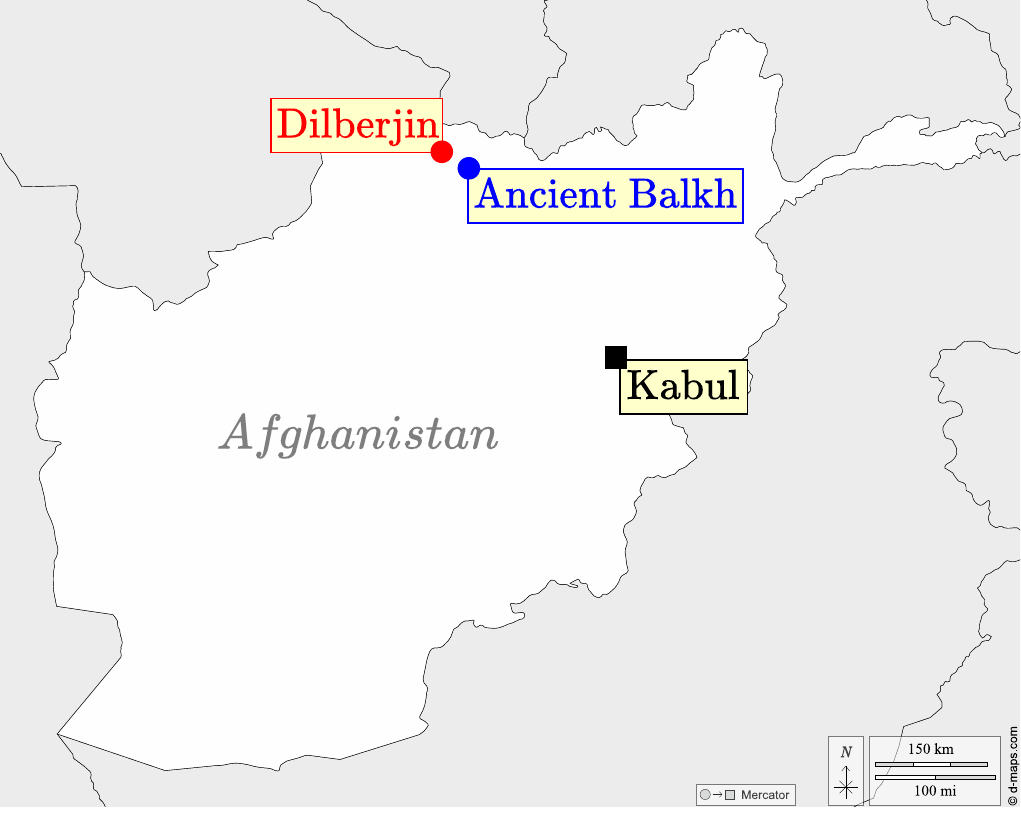}}& \includegraphics[width=0.16\linewidth, cfbox=blue 1pt]{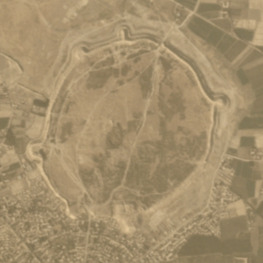}  & \includegraphics[width=0.16\linewidth, cfbox=blue 1pt]{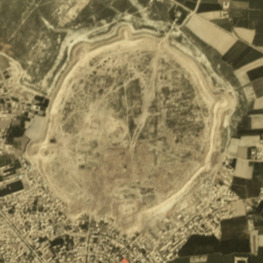}& \includegraphics[width=0.16\linewidth, cfbox=blue 1pt]{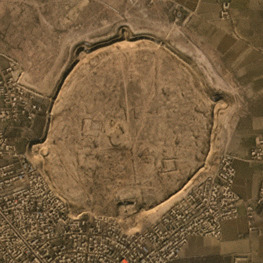}\\
         & \includegraphics[width=0.16\linewidth, cfbox=red 1pt]{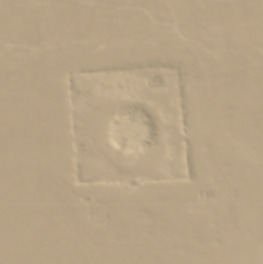}  & \includegraphics[width=0.16\linewidth, cfbox=red 1pt]{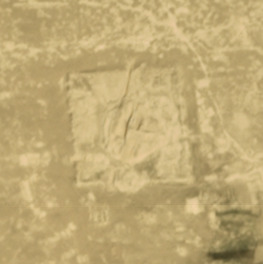}& \includegraphics[width=0.16\linewidth, cfbox=red 1pt]{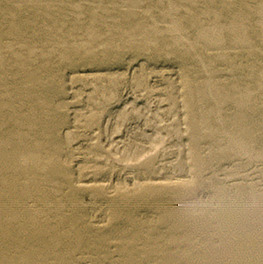}\\
         (c) Map of Afghanistan & (d) Oct. 2017 & (e) Apr. 2020 & (f) Jan. 2023
     \end{tabular}
     }
    \caption{\textbf{DAFA Looted Sites (\datasetname)} contains monthly satellite image time series (SITS) of Afghan archaeological sites acquired between 2016 and 2023. We show the location of preserved (a) and looted (b) sites, adding strong random noise to their coordinates to prevent misuse of the data. Test sites are marked with a star ($\star$). We also show images (d-f) from two sites (c): {\color{blue}Ancient Balkh} (in blue, top row) has been preserved from looting~\cite{unesco2004balkh}, while {\color{red}Dilberjin} (in red, bottom row) suffered irreparable damage.% between 2019 and 2021~\cite{follorou2023en}.
    }\vspace{-1em}
    \label{fig:teaser}
\end{figure}

\section{Introduction}

Between 2003 and 2013, numerous archaeological studies were conducted in Afghanistan resulting in the identification of over a thousand sites through terrestrial surveys and satellite imagery, in particular in the northern part of the country (see Figures~\hyperref[fig:teaser]{\ref*{fig:teaser}a} \&~\hyperref[fig:teaser]{\ref*{fig:teaser}b}). However, the progressive deterioration of security in Afghanistan posed challenges for continued field research and it has remained relatively unexplored by archaeologists since 2015. In 2022, the French archaeological delegation in Afghanistan (DAFA) identified the first instance of looting at the Dilberjin site~\cite{follorou2023en} (Figure~\hyperref[fig:teaser]{\ref*{fig:teaser}f}). Leveraging their extensive knowledge of the terrain, and together with the Iconem startup, archaeologists started documenting widespread pillaging affecting more than a hundred major sites. Given the scale of this phenomenon, an automated approach becomes necessary, and would help identifying more quickly signs of looting. While some research on detecting looted archaeological sites exists (see Section~\ref{sec:rw_archaeo}), it remains limited and most existing publications do not release any data. This lack of public datasets makes it harder for machine learning researchers unfamiliar with archaeology or remote sensing data to contribute their knowledge and skill despite the theoretical and practical interest of the problem. 

We introduce the DAFA Looted Sites dataset (\datasetname), a dataset containing 55,480 images acquired monthly over 8 years, from 2016 to 2023, across 675 Afghan archaeological sites, including 135 sites looted during this period. The dataset is organized into satellite image time series (SITS) and we distinguish between `looted' and `preserved' sites (see Figure~\ref{fig:teaser}). \datasetname~is the first open-access dataset to make it possible for SITS classification methods to evaluate on the looting detection task. Note that looting detection is significantly different from most change detection tasks, since looting can appear as a relatively subtle appearance change, while other significant changes, e.g., in the vegetation, are not relevant.

The paper is structured as follows. In Section~\ref{sec:rw}, we review related works and datasets for archaeological looting detection and SITS classification. In Section~\ref{sec:dataset}, we describe our dataset and its curation. In Section~\ref{sec:benchmark}, we present diverse baseline methods and report their performance. Finally, we discuss the limitations of \datasetname~in Section~\ref{sec:limitations}.
\section{Related Work}\label{sec:rw}

We first provide an overview of the archaeological looting detection literature (Section~\ref{sec:rw_archaeo}) and then give a broad outline of satellite image time series (SITS) classification (Section~\ref{sec:rw_sits}).

\subsection{Archaeological looting detection}\label{sec:rw_archaeo}

\paragraph{SITS-based visual detection of archaeological looting.} Aerial and/or satellite imagery offers a means to manually monitor archaeological sites, complementing time-consuming, expensive and sometimes hazardous ground truth surveys~\cite{parcak2007satellite, contreras2010utility}. Furthermore, human expeditions are impractical in several countries due to political and military restrictions~\cite{coluzzi2010satellite}. Multi-temporal satellite imagery allows archaeologists to identify looting patterns by comparing successive images~\cite{thomas2006recent, parcak2007satellite}. The literature on using remotely sensed data to manually detect potential looting of archaeological sites is rich \cite{stone2008patterns, stone2015update, kennedy2009aerial, thomas2006recent, thomas2008archaeological, parcak2009satellite, parcak2016satellite, aaas2014ancient, casana2014satellite, casana2015satellite, casana2017satellite, agapiou2020detecting, zaina2021multi, tapete2016looting, agapiou2017optical, abate2023aerial}. Early methods relied on raw images for direct visual assessment of damage, without any image enhancement or data processing. For example, Stone~\cite{stone2008patterns, stone2015update} used raw SITS to detect looting holes in Iraq, while Kennedy et al.~\cite{kennedy2009aerial} assessed site destruction by bulldozers in Jordan using aerial image time series. Several publications~\cite{aaas2014ancient, casana2014satellite, casana2015satellite, casana2017satellite} focus on Syrian sites located in conflict zones, visually comparing pre- and post-conflict high-resolution satellite images to identify looting, quantify damage, and assess the timing of these events. More recent work confirms the potential of visual inspection of SITS to identify looting on Syrian sites~\cite{agapiou2020detecting} or other damaging practices (ploughing, building/road/canalization constructions) in Iran~\cite{zaina2021multi}. Another line of work processes the images before visually analyzing them. Tapete et al.~\cite{tapete2016looting} apply a Gaussian filter to the images before ratioing consecutive image pairs in order to enhance morphological changes. This method is used on synthetic aperture radar (SAR) images to detect looting holes or marks on Syrian sites. To ease the process of visual interpretation of the images, Agapiou et al.~\cite{agapiou2017optical} perform photo-enhancing operations, playing on the contrast, the brightness and the histogram of Google Earth multi-temporal images of Cyprus, in order to visually identify looting marks. Finally, Abate et al.~\cite{abate2023aerial} compare multi-temporal orthophotographies acquired by unmanned aerial vehicle (UAV) to the output of the maximum autocorrelation factor/multivariate alteration detection (MAF/MAD) transformations~\cite{nielsen1998multivariate}, commonly used in change detection studies.

\paragraph{Automatic looting detection.} In order to speed up the detection process and allow for a fast response to potential threats, automatic identification methods have also been developed. A simple approach is to use the thresholded difference image between bi-temporal satellite acquisitions in order to reveal looting marks. Rayne et al.~\cite{rayne2020detecting} apply this method on Sentinel-2 images a year apart to detect looting pits in Lybia and Egypt. Castilla et al.~\cite{castilla2009land} compute change maps from bi-temporal satellite image pairs as the sum of their robust difference~\cite{castilla2009land} (\textit{i.e.}, difference in brightness) and the difference between their Gabor feature maps~\cite{manjunath1996texture, han2007rotation} (\textit{i.e.}, difference of texture). This method is applied to pre/post-disaster image pairs to identify sites damaged by terrorists in Syria and Iraq. Bowen et al.~\cite{bowen2017algorithmic} localize looting pits in Egypt with a tree-based classifier using SIFT~\cite{lowe1999object}, SURF~\cite{bay2006surf} and HOG~\cite{dalal2005histograms} descriptors on mono-temporal mid-resolution satellite images. A series of work~\cite{lasaponara2018space, masini2021remote, masini2020recent} investigates a clustering approach on spatial auto-correlation features for the detection of looting holes or pits with bi-temporal Syrian and Peruvian Google Earth images. El Hajj~\cite{el2021interferometric} classifies image patches to reveal whether they contain looting or destruction instances using an ensemble model composed of a random forest~\cite{ho1995random}, an AdaBoost~\cite{freund1997decision} classifier and a SMOTEBoost~\cite{chawla2003smoteboost} classifier. Payntar~\cite{payntar2023multi} characterizes potential destruction of archaeological sites by the change in the number of different land cover classes in a given neighborhood. Quinquennially Landsat images of Peru from 1985 to 2020 are segmented independently with a random forest to obtain the land cover maps. Finally, mask R-CNNs~\cite{he2017mask} have recently been applied for the detection of looting holes on mono-temporal UAV data of several areas across the globe~\cite{altaweel2024monitoring}. Recent methods are trained and evaluated on datasets that, in contrast to \datasetname, are not available in open-access and/or not multi-temporal, as reported in Table~\ref{tab:datasets}.

\begin{table}[t!]
    \centering
    \caption{\textbf{Aerial and satellite image datasets for archaeological site looting detection.} In Masini et al.~\cite{masini2020recent}, images were acquired with Google Earth with varying spatial resolution. El Hajj~\cite{el2021interferometric} does not provide open access to their dataset, but it can theoretically be recreated openly since (i) locations, images and labels come from open-source data, and (ii) the pre-processing steps are detailed in the paper.} 
    \resizebox{\linewidth}{!}{
    \begin{tabular}{lccclllc}
        \toprule
        & Open- & Multi- & Spatial & Temporal  & \multirow{2}{*}{Sensor} & \multirow{2}{*}{Location} & Number\\
        & access & temporal & resolution & resolution & & & of sites \\
        \midrule
        Masini et al. (2020)~\cite{masini2020recent} & \xmark & \cmark & Varying & Yearly & Satellite & Syria & 2\\
        El Hajj (2021)~\cite{el2021interferometric} & \xmark & \xmark & 15m/px & --- & Satellite & Syria and Iraq & 9\\
        Payntar (2023)~\cite{payntar2023multi} & \xmark & \cmark & 30m/px & Quinquennially & Satellite & Peru & 477\\
        Altaweel et al. (2024)~\cite{altaweel2024monitoring} & \cmark & \xmark & 3cm/px & --- & UAV & Worldwide & 95\\
        \midrule
        \textbf{\datasetname~(ours)} & \cmark & \cmark & 3m/px & Monthly & Satellite & Afghanistan & 675\\
        \bottomrule
    \end{tabular}
    }
    \label{tab:datasets}
\end{table}

\subsection{Satellite image time series classification}\label{sec:rw_sits}

\paragraph{Satellite image time series datasets.} Many SITS datasets have been created in recent years for applications to crop-type mapping~\cite{russwurm2018multi, russwurm2019breizhcrops, weikmann2021timesen2crop, garnot2021panoptic, kondmann2021denethor, tarasiou2022context}, wildfire spread prediction~\cite{gerard2023wildfirespreadts}, tree species identification~\cite{astruc2024omnisat}, wilderness mapping~\cite{ekim2023mapinwild}, change detection~\cite{toker2022dynamicearthnet, van2021multi, verma2021qfabric}, land-cover mapping~\cite{garioud2024flair, wenger2022multisenge}, low-to-high resolution knowledge transfer~\cite{li2022outcome}, cloud removal~\cite{ebel2022sen12ms}, and electricity access detection~\cite{ma2021outcome}. \datasetname~is the first open-access dataset for the detection of looted archaeological sites. Compared to other STIS classification datasets and applications, our benchmark presents several important specificities. First, the changes related to looting that we aim at detecting can be extremely subtle and hardly visible, while other very important but irrelevant changes are frequent. Second, because we target regular monitoring at limited cost, our dataset is built from images of limited spatial resolution but available freely at high and uniform temporal resolution. Third, because of the limited number of archaeological sites, and particularly looted archaeological sites, training data is necessarily small and very imbalanced.

\paragraph{Satellite image time series classification approaches.} Existing techniques for satellite image time series classification include pixel-wise and whole-image methods. Pixel-wise methods~\cite{pelletier2019temporal, interdonato2019duplo, russwurm2020self, garnot2020lightweight, vincent2023pixel} process each pixel of the SITS independently, discarding spatial information. In practice, these methods are outperformed by whole-image approaches, like PSE+LTAE~\cite{garnot2020satellite} or TSViT~\cite{tarasiou2023vits}. The former encodes a set of pixels before processing the temporal dimension with an attention mechanism while the latter is a fully spatio-temporal attention-based approach. Closely related to our topic are segmentation methods for SITS like 3D-Unet~\cite{rustowicz2019semantic} or UTAE~\cite{garnot2021panoptic} which are U-Net~\cite{ronneberger2015u} based architecture designed to process the temporal dimension. Very recently, several works~\cite{cong2022satmae, fuller2022satvit, reed2023scale, xiong2024neural, astruc2024omnisat, li2024s2mae, noman2024rethinking, guo2024skysense, li2024masked} have aimed at developing a remote sensing "foundation model", introducing large models pretrained on millions of remotely sensed images, often from different sensors, modalities or spatial and temporal resolutions. These pretrained models have shown good performance on classification and segmentation downstream tasks. We evaluate both pixel-wise and whole-image methods, and demonstrate that foundation models can significantly boost performance. 
\section{Dataset}\label{sec:dataset}

\begin{figure}[t!]
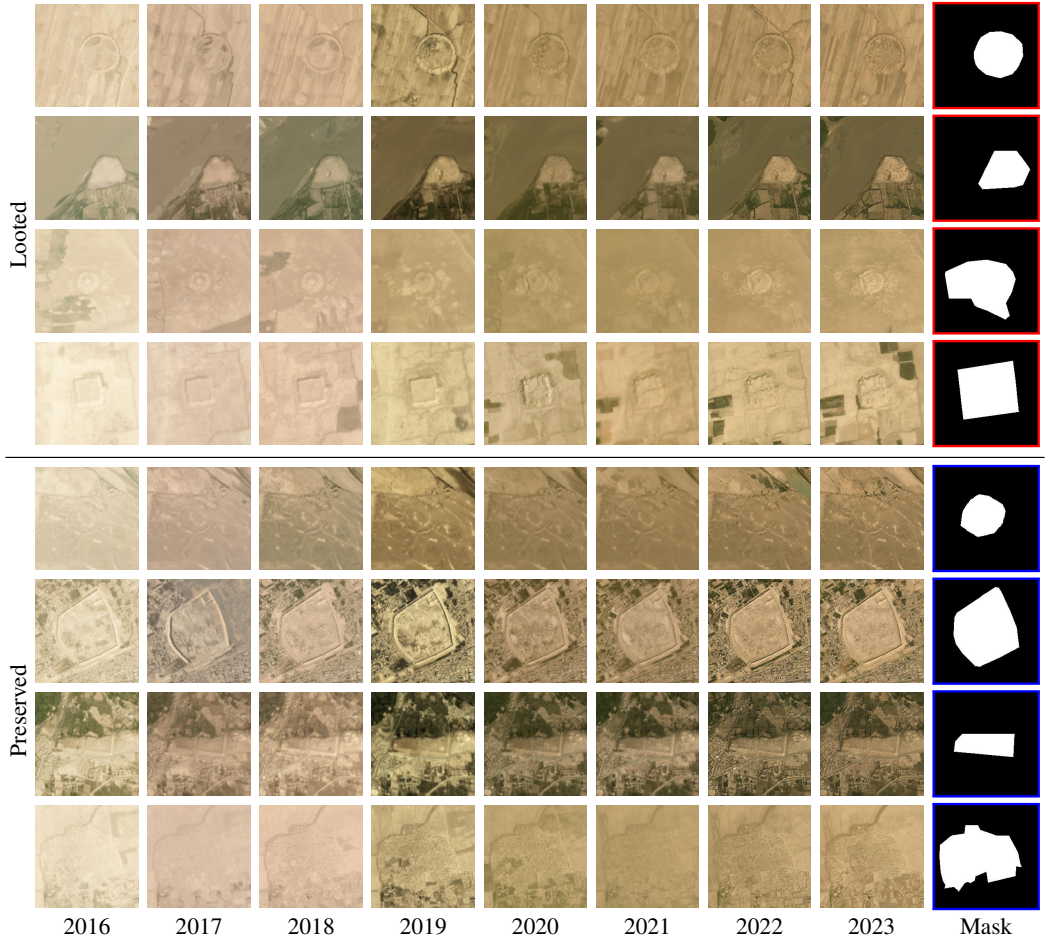

    \renewcommand{\arraystretch}{1}
    \setlength\tabcolsep{2pt}
    \centering
    \resizebox{\linewidth}{!}{
    \begin{tabular}{cccccccccc}
        & \examplesits{looted}{7}{red}
        \multirow{2}{*}{\rotatebox[origin=c]{90}{Looted}} & \examplesits{looted}{78}{red}
        & \examplesits{looted}{115}{red}
        & \examplesits{looted}{123}{red}
        \midrule
        & \examplesits{preserved}{17}{blue}
        \multirow{2}{*}{\rotatebox[origin=c]{90}{Preserved}} & \examplesits{preserved}{133}{blue}
        & \examplesits{preserved}{221}{blue}
        & \examplesits{preserved}{420}{blue}
        & 2016 & 2017 & 2018 & 2019 & 2020 & 2021 & 2022 & 2023 & Mask\\
    \end{tabular}
    }
    \caption{\textbf{Examples of time series and coarse location masks.} For each archaeological site, we show the September image for each year from 2016 to 2023 and the corresponding coarse location mask. The 4 top rows show {\color{red}looted} sites (red squares) and the 4 bottom rows show {\color{blue}preserved} sites (blue squares).}
    \label{fig:examples}
\end{figure}

\paragraph{Sites identification and characterization.} A significant number of Afghan archaeological sites are cataloged in the "Archaeological gazetteer of Afghanistan"~\cite{ball1982archaeological} or documented by the DAFA. Because ground surveys are impractical since 2015 in Afghanistan, our team of expert archaeologists navigated through online platforms of high resolution satellite imagery (Google Earth, ESRI and Bing) in order to enrich the list with new sites and label them into the `looted' and `preserved' categories, depending on whether it has been damaged by malevolent human activities prior to 2023 or not. In total, 986 Afghan archaeological sites have been identified. We have gathered monthly SITS from January 2016 to December 2023 for each location. The source of our images is the Planet Basemaps product\footnote{\url{https://developers.planet.com/docs/basemaps/}} from Planet Labs~\cite{planet2024planet}. Images can be freely downloaded, via the "Planet education and research program", but since we do not want to release geographical locations to avoid misuse, we will instead redistribute modified versions of the images under a CC BY-NC 4.0 license\footnote{\url{https://assets.planet.com/docs/ToS_EducationAndResearch.pdf}}. In addition to the raw acquisitions, Planet applies a proprietary post-processing algorithm in order to minimize the effects of clouds, haze and other image variability~\cite{planet2019planet}. Planet images have an alpha channel indicating areas where there is no image data available. We remove all images with any pixel for which it is the case. This results in some missing time stamps, but the filtered dataset still has a median of 94 dates per time series out of 96 possible dates. Each image of the dataset covers a 1 km² area centered on an archaeological site. Images have 3 channels (RGB) with 266$\times$266 size and a ground sampling distance (GSD) of 3 meters per pixel. 

A 1 km²-square patch centered on a site may cover other archaeological sites in varying states of preservation. Since most sites have a relatively small extent, the majority of the pixels may not provide informative data about potential looting, may contain different archaeological sites, or may contain information that is correlated with looting, such as the presence of roads. For these reasons, leveraging the most recent high-resolution Google Earth image available to date, we manually annotated a binary mask coarsely detouring the sites. A high fraction of the preserved sites could not be delineated with sufficient confidence and was eventually discarded. Several reasons may explain why a site is not visible on a SITS, for example the site may be too small to show on mid-resolution imagery or completely buried under sand or earth. The final number of sites for which a coarse location mask is available is 675 (135 looted, 540 preserved), for a total of 55,480 satellite images. We show some examples of SITS from \datasetname~along with their coarse location mask in Figure~\ref{fig:examples}.

\paragraph{Test set definition.} As can be seen in Figures~\hyperref[fig:teaser]{\ref*{fig:teaser}a} \&~\hyperref[fig:teaser]{\ref*{fig:teaser}b}, looted archaeological sites are mainly located in the northern region of Afghanistan. However, we do not want methods evaluated on our dataset to bypass the looting detection task and learn to locate the sites thanks to geographical cues. For this reason, we ensure that (i) the maximum distance between two test sites is less than 200 km (the maximum distance between two sites in the dataset is more than 1000 km), and that (ii) the maximum distance between two sites of opposite class in the test set is smaller than 40 km (the average is 7.5 km and median 5.5 km). To limit other forms of geographic bias, we also ensure that no site in the test set is closer than 1 km from a site in the rest of the dataset.

\begin{table}[t]
    \centering
    \caption{\textbf{Distribution of areas for each class.} Classifying all sites with a 3-hectare threshold leads to a mean accuracy of 79.4\%, showing the need for a balanced test set to prevent the area bias.}
    \begin{tabular}{lcc}
        \toprule
        Surface (ha) & Looted (\%) & Preserved (\%)\\
        \midrule
        $0\ \ <$...$<1$   & 0    & 16.1\\
        $1\ \ <$...$<2$   & 2.2  & 34.8\\
        $2\ \ <$...$<3$   & 12.6 & 22.6\\
        $3\ \ <$...$<5$   & 22.2 & 14.4\\
        $5\ \ <$...$<10$  & 34.1 & 8.5 \\
        $10   <$...$<100$ & 28.9 & 3.5 \\
        \bottomrule
    \end{tabular}
    \label{tab:surface_distrib}
\end{table}

We also found there was a potential bias in the sites areas, which we can estimate using the coarse location masks. The area distribution for both classes is reported in Table~\ref{tab:surface_distrib}. We observe significantly different distributions between looted and preserved sites, where a simple 3-hectare threshold results in a mean classification accuracy of 79.4\%. To avoid this bias, we ensure that the test set exhibits similar area distributions between the two classes by selecting the same number of sites within each area bin.

Selection using these criteria leads to a total of 61 sites (26 looted, 35 preserved) which we use as our test set.

\paragraph{Training and validation splits.}  From the remaining 614 sites we define 6 splits. First, we define 5 diverse validation splits of 18 sites each ensuring that all sites in a validation split are further than 1 km from any site out of this split. The remaining 524 sites are always used as training. We evaluate all methods using a 5-fold scheme, using one of the validation split as validation and adding the others to the 524 samples we always use for training.
\section{Benchmark}\label{sec:benchmark}

\subsection{Task and evaluation}\label{sec:problem}

\paragraph{Problem formalization.} 

Let $x$ be an RGB input time series of satellite images consisting of $T$ images of height $H$ and width $W$. Each time series corresponds to a unique archaeological site, whose coarse location is given by a binary mask $m$ of size $H\times W$. Given a pair $(x, m)$, the task is to predict a label $y$ in $\{0,1\}$ indicating whether the corresponding site is preserved ($y=0$) or has been looted ($y=1$). Except when stated otherwise, we use as input to all methods the element-wise multiplication of $x$ and $m$, which prevents them from leveraging information contained outside of the coarse location mask which may be correlated with looting, for example the proximity of a road.

\paragraph{Metrics.} We evaluate the classification performance of the baselines using a set of metrics commonly used in the binary classification literature: the overall accuracy (OA), the F1-score (F1), and the area under the ROC curve (AUROC). Additionally, we use the false alarm rate (FAR), also known as the false positive rate and common in the looting detection literature~\cite{lasaponara2014investigating, lasaponara2018space, masini2020recent, masini2021remote}. A low number of false positives is indeed desired because it limits the number of costly and time-consuming verifications and potential ground surveys on sites flagged as looted. 

\subsection{Baselines}

We evaluated three types of methods: single frame methods, pixel-wise multi-frame methods and whole-image multi-frame methods. We summarize their key characteristics and differences in Table~\ref{tab:baselines} of the appendix.

\paragraph{Single-frame methods.} We know looted sites have been damaged prior to 2023 so that looting marks should appear on all 2023 images. Based on this observation, we train single-frame methods only looking at 2023 images and classifying them between ‘looted’ and ‘preserved’ depending on the time series they are taken from. At inference, all the 2023 images of a given time series are input to the model separately, and the most predicted label out of the 12 predictions is selected for the corresponding site. We use as baselines ResNets~\cite{he2016deep} of various sizes trained from scratch and several frozen pre-trained foundation models: SatMAE~\cite{cong2022satmae}, Scale-MAE~\cite{reed2023scale} and DOFA~\cite{xiong2024neural} on top of which we train a linear classification head. They are visual transformers (ViT)~\cite{dosovitskiy2020image} pretrained following the mask-autoencoding (MAE)~\cite{he2022masked} strategy. They also have their own specificities, including the dataset they have been trained on. In particular, DOFA is pretrained on more than 8 million satellite images from different sensors, modalities and spatial resolutions. 

\paragraph{Pixel-wise multi-frame methods.} We can also view \datasetname~as a set of pixel time series and evaluate several methods designed for pixel-wise satellite image time series classification including DuPLo~\cite{interdonato2019duplo}, TempCNN~\cite{pelletier2019temporal}, a self-attention approach referred to as Transformer~\cite{russwurm2020self}, and LTAE~\cite{garnot2020lightweight}. Here, only pixels $x_{i,j}$ located inside the coarse location mask (\textit{i.e.}, $m_{i,j}=1$), are used at training and inference for a given SITS $x$. At inference, we select the most predicted label between ‘looted’ and ‘preserved’ among in-mask pixels as the prediction for a given SITS.

\paragraph{Whole-image multi-frame methods.} We evaluate several methods designed for the SITS classification task as defined in Section~\ref{sec:problem}: PSE~\cite{garnot2020satellite}+LTAE~\cite{garnot2020lightweight}, TSViT~\cite{tarasiou2023vits}, SatMAE~\cite{cong2022satmae}+LTAE~\cite{garnot2020lightweight}, Scale-MAE~\cite{reed2023scale}+LTAE~\cite{garnot2020lightweight} and DOFA~\cite{xiong2024neural}+LTAE~\cite{garnot2020lightweight}. LTAE is a temporal self-attention network that we apply to series of features extracted either with a pixel-set encoder (PSE) or with a frozen pretrained foundation model (SatMAE, Scale-MAE or DOFA). The Temporo-Spatial Vision Transformer (TSViT) has a fully-attentional architecture, processing the tokens first temporally then spatially. We train a small version of TSViT from scratch on \datasetname~with a classification head.

Additionally, we considered seeing the problem as a segmentation one to evaluate a wider range of methods. In this case, we do not mask the images and first train a method to segment the time series at the pixel level into three classes: `looted site' (in-mask pixels for SITS of looted sites), `preserved site' (in-mask pixels for SITS of preserved sites) and `not a site' (out-mask pixels).  At inference, the classification prediction corresponds to the majority class at the pixel-level inside the coarse location mask among `looted' and `preserved'. This enables us to evaluate two small versions of TSViT~\cite{tarasiou2023vits} (with a segmentation head) and UTAE~\cite{garnot2021panoptic} trained from scratch. The U-Net with Temporal Attention Encoder (UTAE) consists of a U-Net architecture where a temporal attention mechanism squeezes the temporal dimension before the decoding branch.

\begin{table}[ht]
    \centering
    \renewcommand{\arraystretch}{1.05}
    \caption{\textbf{Classification performance.} We evaluate several methods trained on \datasetname~and distinguish between methods that process a single image at a time (mono-temporal) and methods receiving time series as input (multi-temporal). Multi-temporal methods can be pixel-wise or whole-image based. We indicate with a star (\textsuperscript{$\star$}) methods that have been pretrained on another dataset beforehand. Best scores are highlighted in \textbf{bold} and second best are \underline{underlined}. We show the standard deviations over the 5 folds in (parenthesis) and report the number of trainable parameters for each baseline.}
    \resizebox{\linewidth}{!}{
    \begin{tabular}{lrcccc}
        \toprule
        Method & \#param (x1000) & OA$\uparrow$ & F1$\uparrow$ & AUROC$\uparrow$ & FAR$\downarrow$\\
        \midrule
        \textit{Single-frame methods} \\
        ~~~~~~~~~~~~ResNet20~\cite{he2016deep} & 269.2 & 54.7 (8.9) & 54.5 (17.1) & 75.3 (3.1) & 54.9 (34.5)\\
        ~~~~~~~~~~~~ResNet18~\cite{he2016deep} & 11,177.5 & 71.8 (2.6) & 64.1 (5.4) & 84.5 (1.5) & 19.4 (3.3)\\
        ~~~~~~~~~~~~ResNet34~\cite{he2016deep} & 21,285.7 & 74.1 (3.2) & \underline{68.9} (6.3) & \underline{85.2} (1.7) & 22.3 (8.0)\\
        ~~~~~~~~~~~~SatMAE\textsuperscript{$\star$}~\cite{cong2022satmae} & 2.1 & 63.6 (0.7) & 41.9 (0.4) & 75.3 (0.2) & \underline{12.0 }(1.1)\\
        ~~~~~~~~~~~~Scale-MAE\textsuperscript{$\star$}~\cite{reed2023scale} & 2.1 & 62.6 (0.7) & 39.3 (1.9) & 76.0 (0.3) & \underline{12.0} (1.1)\\
        ~~~~~~~~~~~~DOFA\textsuperscript{$\star$}~\cite{xiong2024neural} & 1.5 & \underline{76.7} (2.8) & 67.0 (4.2) & 84.0 (1.4) & \textbf{7.4} (2.3) \\
        \midrule
        \textit{Multi-frame methods} \\
        ~~~~~~\textit{Pixel-wise methods} \\
        ~~~~~~~~~~~~DuPLo~\cite{interdonato2019duplo} & 86.8 & 52.1 (2.8) & 50.4 (4.9) & 50.9 (3.7) & 52.0 (7.8)\\
        ~~~~~~~~~~~~TempCNN~\cite{pelletier2019temporal} & 28.5 & 55.7 (3.4) & 44.2 (9.7) & 58.8 (1.8) & 34.9 (9.8)\\
        ~~~~~~~~~~~~Transformer~\cite{russwurm2020self} & 38.5 & 56.4 (3.7) & 63.5 (3.2) & 62.7 (4.1) & 68.0 (10.0)\\
        ~~~~~~~~~~~~LTAE~\cite{garnot2020lightweight} & 32.2 & 52.5 (7.8) & 58.0 (4.6) & 62.0 (8.5) & 65.7 (18.8) \\
        ~~~~~~\textit{Whole-image methods} \\
        ~~~~~~~~~~~~PSE~\cite{garnot2020satellite}+LTAE~\cite{garnot2020lightweight} & 34.0 & 55.1 (9.8) & 47.7 (6.2) & 59.5 (6.3) & 39.4 (19.4)\\
        ~~~~~~~~~~~~UTAE~\cite{garnot2021panoptic} & 68.9 & 62.0 (3.5) & 58.9 (2.3) & 64.5 (4.5) & 39.4 (8.6)\\
        ~~~~~~~~~~~~TSViT~\cite{tarasiou2023vits} (cls. head) & 236.9 & 64.3 (1.2) & 53.0 (3.7) & 70.8 (2.3) & 23.4 (4.9) \\
        ~~~~~~~~~~~~TSViT~\cite{tarasiou2023vits} (seg. head) & 237.4 & 64.6 (3.5) & 60.2 (7.1) & 69.6 (4.2) & 35.4 (6.9) \\
        ~~~~~~~~~~~~SatMAE\textsuperscript{$\star$}~\cite{cong2022satmae}+LTAE~\cite{garnot2020satellite} & 1,627.9 & 67.9 (4.7) & 64.7 (4.0) & 75.2 (3.7) & 33.1 (11.1)\\
        ~~~~~~~~~~~~Scale-MAE\textsuperscript{$\star$}~\cite{reed2023scale}+LTAE~\cite{garnot2020satellite} & 1,627.9 & 68.5 (2.4) & 56.4 (7.7) & 77.6 (0.8) & 17.1 (4.4)\\
        ~~~~~~~~~~~~DOFA\textsuperscript{$\star$}~\cite{xiong2024neural}+LTAE~\cite{garnot2020satellite} & 926.1 & \textbf{78.7} (2.3) & \textbf{74.9} (3.5) & \textbf{87.1} (3.0) & 18.9 (6.9)\\
        \bottomrule
    \end{tabular}
    }
    \label{tab:results}
\end{table}

\paragraph{Implementation details.}\label{sec:implem} 
We trained all methods on a single NVIDIA GeForce RTX 2080 Ti or NVIDIA V100 GPU. All methods are trained using a binary-cross entropy loss and the AdamW optimizer~\cite{loshchilov2017decoupled}, except for the segmentation methods that have been trained using a regular cross-entropy loss. We use the implementations of DuPLo and TempCNN available in Transformer's official public repository\footnote{\url{https://github.com/MarcCoru/crop-type-mapping}} and the official implementation for all other methods. We use random resize crop, random rotate and random flip as data augmentation for whole-image methods. For whole-image time series based approaches, 24 dates (3 per year) are randomly sampled out of the 96 available at training time and the whole time stamps are used at inference. Additional details on training and architecture configurations can be found in the appendix.

\subsection{Results}

We report the performance of all evaluated baselines in Table~\ref{tab:results} and make three key observations.
(i) First, leveraging the DOFA foundation model outperforms other methods. Among single-frame methods, DOFA surpasses ResNet20 and ResNet18 trained from scratch, as well as the frozen SatMAE and Scale-MAE models, and is on par with the larger ResNet34 trained from scratch but with a significantly lower FAR. In the multi-frame category, DOFA+LTAE clearly outperforms PSE+LTAE, SatMAE+LTAE, and Scale-MAE+LTAE, delivering our best overall performance. Note that, altough DOFA has not been trained on Planet imagery and at its particular spatial resolution, the pre-training set of DOFA includes SatlasPretrain~\cite{bastani2023satlaspretrain} that contains images of Afghanistan in the SAR modality. DOFA (111M parameters) also has multiple orders of magnitude more parameters than ResNet20 (0.27M parameters) for example. (ii) Second, the temporal information benefits from a specific treatment, beyond a simple voting aggregation strategy: DOFA+LTEA outperforms the single-frame DOFA baseline by +7.9pt in F1 score, +3.1pt in AUROC, and +2.0pt in OA. Note that this comes at the cost of an increased false alarm rate. This observation is consistent among all evaluated foundation models, with SatMAE+LTAE and Scale-MAE+LTAE outperforming their single-frame counterparts at the cost of an increased FAR. (iii) Finally, as often with SITS data, pixel-wise methods are significantly outperformed by whole-image approaches, showing the importance of spatial information for the looting detection task. 

\begin{figure}[t!]
    \renewcommand{\arraystretch}{1}
    \setlength\tabcolsep{0pt}
    \centering
    \begin{tabular}{cc}
    \includegraphics[width=0.6\linewidth]{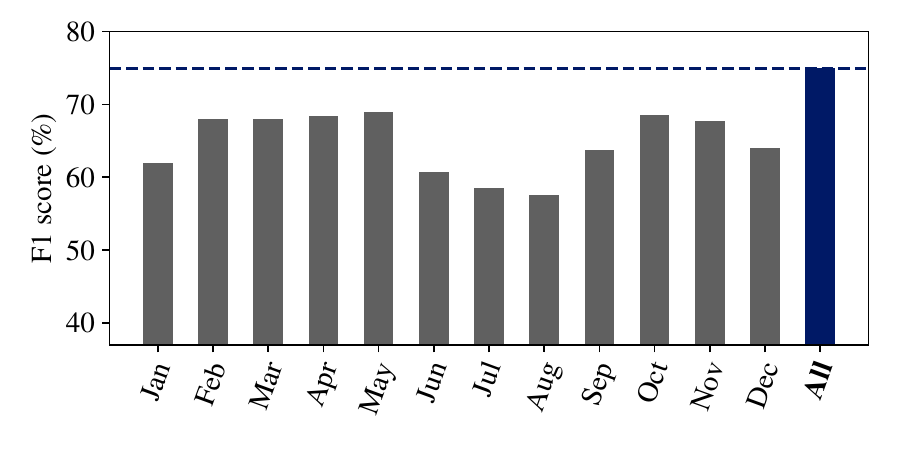} & \includegraphics[width=0.4\linewidth]{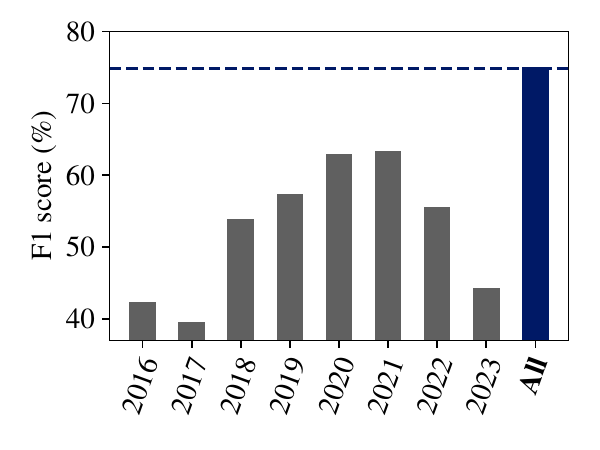} \\
    (a) Month-specific time series & (b) Year-specific time series
    \end{tabular}
    \caption{\textbf{Ablation of DOFA+LTAE.} We evaluate DOFA+LTAE on either month-specific (a) or year-specific (b) sub-time series of \datasetname. In other words, instead of taking all monthly time stamps as input, we only take images of a given month (across all years) or of a given year.}
    \label{fig:ablation}
\end{figure}

\paragraph{Ablation.} We further evaluate whether methods can learn temporal cues from \datasetname~despite the lack of frame-wise annotations. To achieve this, we use our best-performing baseline (DOFA+LTAE) and perform inference on subsets of the time series. We first constructed yearly time series by taking images from a specific month across all available years. As shown in Figure~\hyperref[fig:ablation]{\ref*{fig:ablation}a}, using yearly month-specific sub-series results in performance that is similar or worse compared to using all available time stamps. Notably, inferences made on spring or autumnal time series are comparable to those made on all-month time series in terms of F1 score. This experiment demonstrates (i) that season — and likely vegetation — affects detection performance, and (ii) that the temporal attention mechanism of LTAE can prioritize informative time stamps when using monthly time series. Second, we conducted inference on monthly year-specific time series. A similar conclusion can be drawn from the results shown in Figure~\hyperref[fig:ablation]{\ref*{fig:ablation}b}. We observe a performance peak using the 2020 and 2021 sub-time series, suggesting that looting-related changes (appearance of marks or scars) likely occurred during this period. A further analysis of the contribution of temporal data to performance is available in Section~\ref{sec:temp_analysis} of the appendix.

\section{Limitations and Discussion}\label{sec:limitations}

Our work~comes with several limitations. First, all archaeological sites are located in Afghanistan, a relevant case study for which we have reliable looting annotations. Consequently, most images are of dry and desertic areas. The conclusions drawn on our dataset will likely not hold in regions with dense vegetation, where archaeological sites are hidden beneath the canopy, and other modalities like SAR or LiDAR may be required.

Second, to avoid biases, we chose to discard pixels outside the coarse location mask in most of our baselines. While these pixels may not provide direct evidence of looting, analyzing the surroundings of archaeological sites could offer insights into factors that increase the likelihood of looting. \datasetname~includes a 1 km² area around each site, allowing for such analysis in future research.

Third, although the weak annotations in \datasetname~make it a challenging dataset, the classification task could benefit from temporal annotations of the damage (\textit{i.e.}, the dates when marks appear) and spatial labeling of looting scars at the pixel level. However, higher resolution imagery would likely be necessary for this purpose, but it is not freely accessible and is typically not available at high temporal resolution.

Finally, we only considered supervised baseline approaches, but given the lack of fine-grained annotations, unsupervised change detection methods might be useful for discovering looting events. We believe the variability of our image would make unsupervised approaches challenging, but we hope \datasetname~will encourage investigations of such methods and their potential combination with the available weak annotation labels.

\paragraph{Potential negative societal impact.} The purpose of this dataset is to encourage the development of efficient methods for detecting looted archaeological sites. The primary goal is the automatic and remote monitoring of known sites to preserve the cultural heritage they represent. To prevent misuse by malevolent individuals or organizations, we do not release the GPS coordinates of the sites in \datasetname, added random noise to the point coordinates before plotting the maps in Figures~\hyperref[fig:teaser]{\ref*{fig:teaser}a} \&~\hyperref[fig:teaser]{\ref*{fig:teaser}b} and added random geometric transformations to the SITS we distribute.

\paragraph{Ethical statement.} Following the template of Sandbrook et al. (2021)~\cite{sandbrook2021principles}, we have ensured that \datasetname~complies with the set of ethical principles outlined below:

\textit{(a) Recognize and acknowledge:} We recognize and acknowledge that the release of DAFA-LS can have social impacts. In particular, a looting dataset has additional security considerations, and may cause undue harm to individuals.

\textit{(b) Necessity and proportionality:} Archaeological sites all over the world face natural and anthropological threats. The rich literature about site monitoring confirms it is necessary to ensure their protection as they constitute a cultural heritage. The use of freely available satellite data to flag their potential looting is a proportional action to address this conservation problem. The interest of a satellite monitoring tool is to identify looting activities as quickly as possible in order to inform the country’s authorities, international experts (UNESCO, ICOMOS), and the general public.

\textit{(c) Potential impacts on people:} In a politically unstable country such as Afghanistan, looting detection exposes the locals at risk of unforeseen legal consequences. There exists a risk to privacy, to safety and security of locals, including Human rights and to irresponsible use, and violation of legal compliance.

\textit{(d) Consent from people:} DAFA-LS is built on authorized and official digs and observations made by Philippe Marquis (co-author), as a member of the DAFA (Délégation Archeologique Francaise en Afghanistan). French DAFA archaeologists have been helping local archaeologists since 1922, in answer to a wish from King Amanullah, which was renewed by the President of the Republic, Ashraf Ghani, in the 2010s~\cite{bendezu2018preserver}. The “\textit{Traité d’amitié et de coopération entre la République française et la République islamique d’Afghanistan}”~\cite{france2012loi}, ratified in 2012 for a duration of 20 years, reinforces the legitimacy of DAFA’s presence and action in Afghanistan. At the moment of writing, the \textit{de facto} Taliban regime has not yet reintegrated into the international community of the UN. An institution like the DAFA therefore cannot officially initiate scientific collaborations with its Afghan counterparts. Thus, no Afghan authority can be mentioned at this stage of the work.

\textit{(e) Transparency and accountability:} DAFA-LS is an open-access dataset created from freely available satellite images. All related code is made open-source on the official GitHub repository linked to the dataset. The spatial resolution of the images (> 3m/px) makes it impossible to identify individuals or their personal vehicle. The images represent an area of 1 km² centered on archaeological sites. Therefore, DAFA-LS cannot be used for another purpose than archaeological preservation.

\textit{(f) Peoples' rights and vulnerabilities:} The destruction of cultural heritage is prohibited in Afghanistan and has repeatedly been considered a war crime by the International Criminal Court~\cite{international2021policy}. These acts primarily harm the Afghan people by erasing traces of their past, which are crucial for national cohesion and the country's future. Afghan communities living near archaeological sites could also suffer from the malicious destruction and looting of these sites.

\section{Conclusion}

We release \datasetname, a new dataset for detecting looted archaeological sites. The dataset contains monthly satellite image time series centered on sites in Afghanistan. We use this dataset to evaluate SITS classification approaches, including single-frame methods and pixel-wise or whole-image multi-frame methods. We present a strong baseline that combines the large foundation model DOFA with the LTAE architecture, specifically designed for SITS. We hope this work will encourage the community to improve classification performance on archaeological satellite data, as it is crucial for the preservation of our cultural heritage.
\begin{ack}
The work of Mathieu Aubry was supported by the European Research Council (ERC project DISCOVER, number 101076028). Jean Ponce was supported by the Louis Vuitton/ENS chair on artificial intelligence and the French government under management of Agence Nationale de la Recherche as part of the \textit{Investissements d’avenir} program, reference ANR19-P3IA0001 (PRAIRIE 3IA Institute). This work was granted access to the HPC resources of IDRIS under the allocation 2024-AD011015272 made by GENCI. We thank Titien Bartette, Charlotte Fafet and Loïc Landrieu for their valuable feedbacks, and Guillaume Astruc and Lucas Ventura for their careful proofreading.
\end{ack}

{
\small
\bibliographystyle{abbrv}
\bibliography{egbib}
}
\newpage

\appendix
\section*{Appendix}
We first provide additional visualizations (Section~\ref{sec:visu}) and temporal analysis (Section~\ref{sec:temp_analysis}), and then complete the benchmark with implementations details (Section~\ref{sec:impl}) and further clarifications (Section~\ref{sec:bench}). Our official repository and information on how to download the dataset can be found at \url{https://github.com/ElliotVincent/DAFA-LS}.

\setcounter{table}{0}
\renewcommand{\thetable}{A\arabic{table}}
\setcounter{equation}{0}
\renewcommand{\theequation}{A\arabic{equation}}
\setcounter{figure}{0}
\renewcommand{\thefigure}{A\arabic{figure}}

\section{Additional visualizations}\label{sec:visu}

\subsection{Examples of looting marks}

\begin{figure}[!ht]
    \centering
    \includegraphics[width=0.49\linewidth]{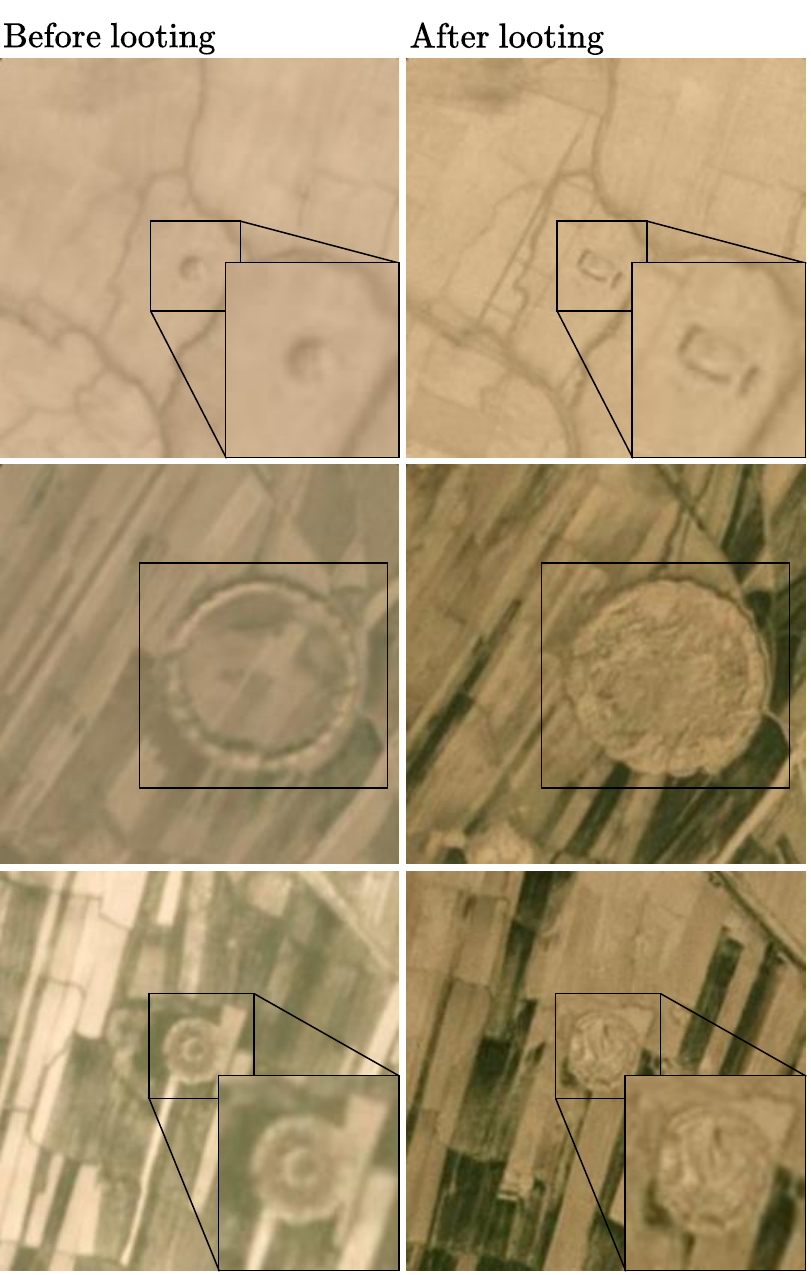}
    \caption{\textbf{Example of visible looting marks.}}
    \label{fig:looting_marks}
\end{figure}

We show in Figure~\ref{fig:looting_marks} three examples of visible looted marks. For the selected sites, we show images before and after the looting, with a zoom on the damaged area. The scars are typical of mechanical looting performed with bulldozers for example.

\subsection{Failure cases}

\begin{figure}[!ht]
    \centering
    \includegraphics[width=\linewidth]{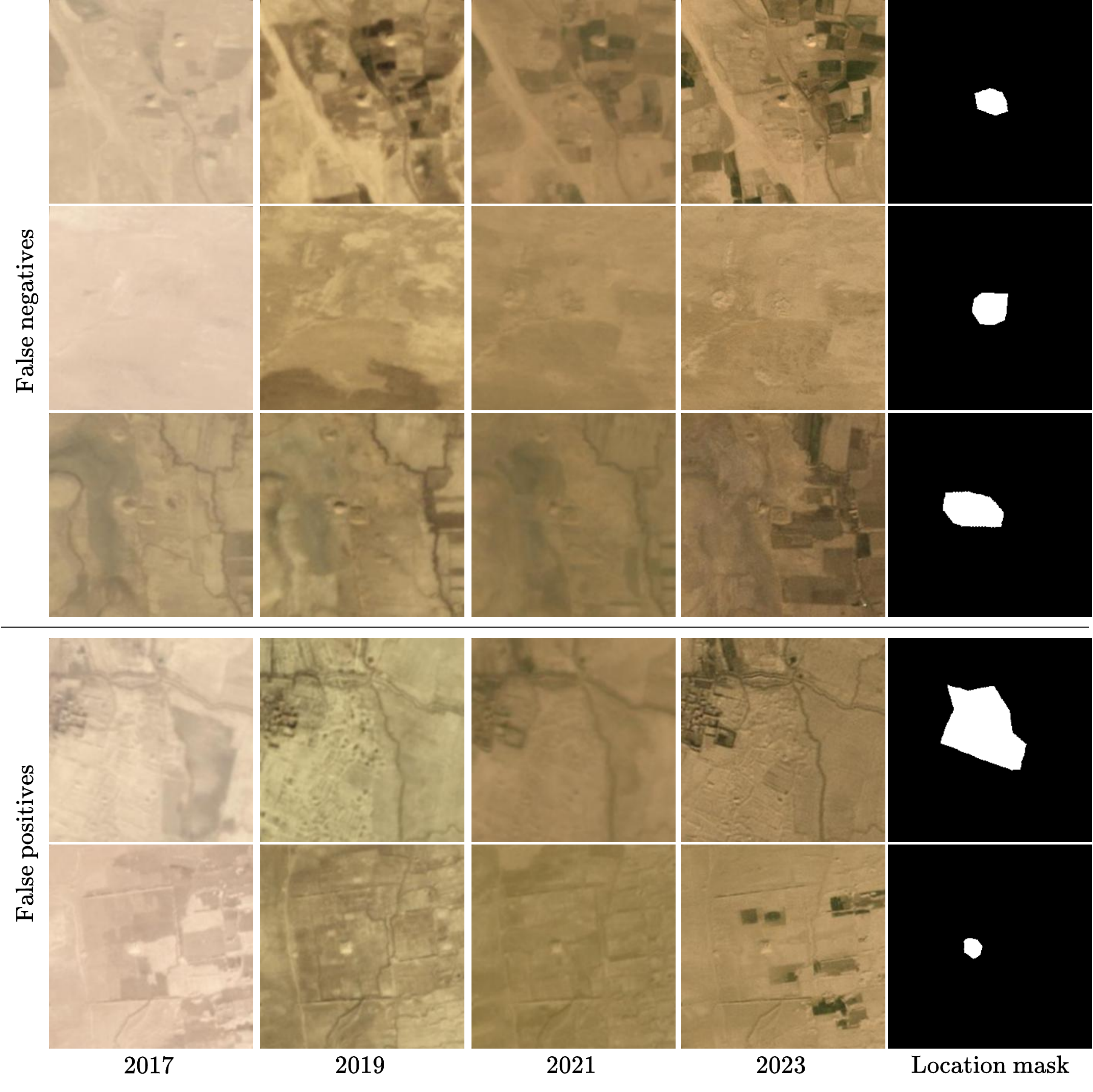}
    \caption{\textbf{Example of failure cases.} We show all the time series for which our best baseline (DOFA+LTAE) predicts the wrong label with a confidence higher than 95\%.}
    \label{fig:failure_cases}
\end{figure}

We report in Figure~\ref{fig:failure_cases} all the time series for which our best baseline (DOFA+LTAE) predicts the wrong label with a confidence higher than 95\%. We note that, for these sites, it is very difficult for a non-expert human eye to identify looting marks if any, or even to clearly see the structure of an archaeological site.

\setcounter{table}{0}
\renewcommand{\thetable}{B\arabic{table}}
\setcounter{equation}{0}
\renewcommand{\theequation}{B\arabic{equation}}
\setcounter{figure}{0}
\renewcommand{\thefigure}{B\arabic{figure}}

\section{Temporal analysis}\label{sec:temp_analysis}

\subsection{Additional ablation}

\begin{figure}[!ht]
    \centering
    \includegraphics[width=0.49\linewidth]{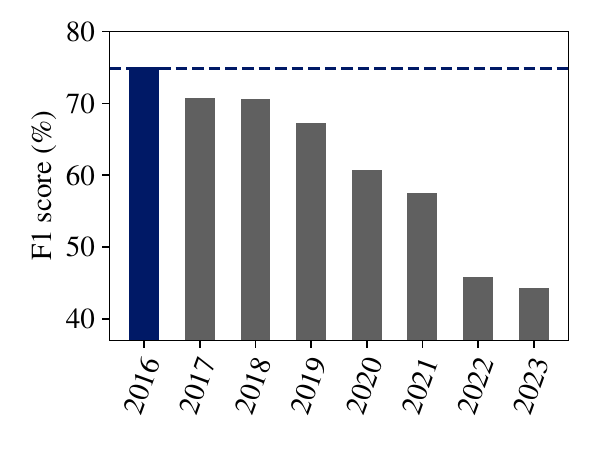}
    \caption{\textbf{Inference with shorter sequence length.} We perform inferences with various temporal range using DOFA+LTAE. We use time series spanning period from 20XX to 2023 with 20XX being the year indicated on the x-axis.}
    \label{fig:various_length}
\end{figure}

We provide in Figure~\ref{fig:various_length} a third ablation, using time series of various lengths at inference for our best performing baseline, DOFA+LTAE. Time series span periods from 20XX to 2023. We observe that the performance consistently decreases as the series becomes shorter.

\subsection{Attention weights}

\begin{figure}[!ht]
    \renewcommand{\arraystretch}{1}
    \setlength\tabcolsep{0pt}
    \centering
    \begin{tabular}{cc}
    \includegraphics[width=0.49\linewidth]{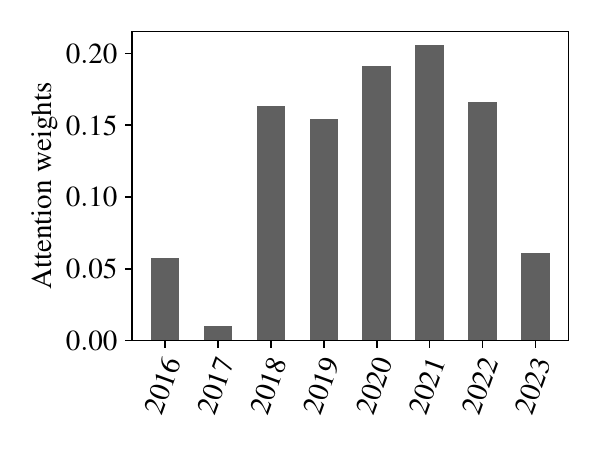} & \includegraphics[width=0.49\linewidth]{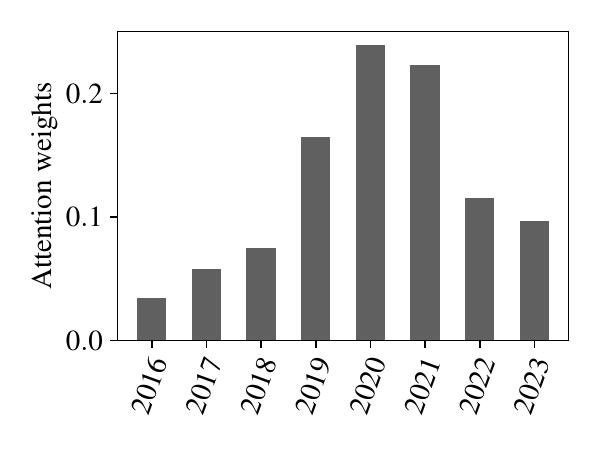}   \\
    (a) Head n°2/8 & (b) Head n°3/8 \\
    \end{tabular}
    \caption{\textbf{Temporal attention.} We report the attention weights for two of the eight heads of DOFA+LTAE. We report the mean of the weights for all looted test sites. We sum all the monthly weights for each year.}
    \label{fig:attention}
\end{figure}

We visualize in Figure~\ref{fig:attention} the temporal attention weights, gathered by year and averaged over all looted test sites, for two attention heads of DOFA+LTAE. We can see that the years 2020 and 2021 draw more attention from the model.

\setcounter{table}{0}
\renewcommand{\thetable}{C\arabic{table}}
\setcounter{equation}{0}
\renewcommand{\theequation}{C\arabic{equation}}
\setcounter{figure}{0}
\renewcommand{\thefigure}{C\arabic{figure}}

\section{Implementation details}\label{sec:impl}

\subsection{Single-frame methods}

We use the PyTorch implementation of ResNet20 for CIFAR-10 by Yerlan Idelbayev\footnote{\url{https://github.com/akamaster/pytorch_resnet_cifar10}}, the torchvision implementation of ResNet18 and ResNet34, and the official PyTorch implementation of SatMAE\footnote{\url{https://github.com/sustainlab-group/SatMAE}}, Scale-MAE\footnote{\url{https://github.com/bair-climate-initiative/scale-mae}} and DOFA\footnote{\url{https://github.com/zhu-xlab/DOFA}}. We use a base version of foundations models when available (DOFA) and a large version otherwise (SatMAE, Scale-MAE). The models are trained with a learning rate of 10$^{-3}$ for 60 epochs and a batch size of 32 (13,380 iterations). ResNet20 is trained from scratch. We train a single linear layer on top of a pretrained frozen DOFA. 

\subsection{Pixel-wise multi-frame methods}

We use the PyTorch implementations of DuPLo and TempCNN available in Transformer’s official public repository\footnote{\url{https://github.com/MarcCoru/crop-type-mappin}} and the PyTorch implementation of LTAE available in UTAE's official public repository\footnote{\url{https://github.com/VSainteuf/utae-paps}}. These models are trained with a learning rate of 10$^{-4}$ for 2 epochs and a batch size of 128 (25,000 iterations). Each batch contains 128 pixel-wise time series, sampled from possibly different SITS. For all these methods, we use smaller versions of the architecture compared to their default setting, taking into account the relative small size of DAFA-LS and to limit over-fitting. The used configurations can be found in our official repository\footnote{\url{https://github.com/ElliotVincent/DAFA-LS}}. 

\subsection{Whole-image multi-frame methods}

We use the official PyTorch implementations of UTAE\footnote{See footnote 4.}and TSViT\footnote{\url{https://github.com/michaeltrs/DeepSatModels}}. These models are trained with a learning rate of 10$^{-4}$ for 100 epochs and a batch size of 4 (14,900 iterations). For PSE+LTAE, SatMAE+LTAE, Scale-MAE+LTAE and DOFA+LTAE, we use SatMAE, Scale-MAE, DOFA and LTAE implementations mentioned above and the official implementation of PSE\footnote{\url{https://github.com/VSainteuf/pytorch-psetae}}. These models are trained with a learning rate of 10$^{-4}$ for 200 epochs and a batch size of 8 (14,900 iterations). For PSE+LTAE, 1024 randomly sampled in-mask pixels are used during training an:q:d all in-mask pixels are used at inference. A majority voting rule is applied at inference to determine the final prediction for a given SITS. For SatMAE+LTAE, Scale-MAE+LTAE and DOFA+LTAE, the backbone is loaded with pretrained weights and frozen during training. We use smaller versions UTAE and TSViT compared to their default setting, taking into account the relative small size of DAFA-LS and to limit over-fitting. The used configurations can be found in our official repository\footnote{See footnote 5}. 

\setcounter{table}{0}
\renewcommand{\thetable}{D\arabic{table}}
\setcounter{equation}{0}
\renewcommand{\theequation}{D\arabic{equation}}
\setcounter{figure}{0}
\renewcommand{\thefigure}{D\arabic{figure}}

\section{Benchmark: additional details}\label{sec:bench}

We detail in Table~\ref{tab:baselines} the main characteristics of the evaluated baselines. In particular, we report the dates available at training time (\textit{Dates used}), the learning task (\textit{Task}), and the inference procedure (\textit{Inference strategy}).

\begin{table}[!h]
    \centering
    \caption{\textbf{Categorization of baseline methods.} We explicit the main characteristics of the different evaluated baseline methods. }
    \renewcommand{\arraystretch}{1.05}
    \resizebox{\linewidth}{!}{
    \begin{tabular}{lllllll}
        \toprule
        \rowcolor{gray!20}Id & Category & Image level & Dates used & Input type & Task & Inference strategy \\
        \midrule
        (i) & Single-frame & Whole-image & 2023 & Image & Classification & 12-month image voting \\
        (ii) & Multi-frame & Pixel-wise & 2016-2023 & SITS & Classification & In-mask pixel voting \\ 
        (iii) & Multi-frame & Whole-image & 2016-2023 & SITS & Segmentation & In-mask pixel voting\\
        (iv) & Multi-frame & Whole-image & 2016-2023 & SITS & Classification & Direct prediction\\
        \midrule
        \rowcolor{gray!20}Id & \multicolumn{6}{l}{Model name} \\
        \midrule
        (i) & \multicolumn{6}{l}{ResNet20/18/34~\cite{he2016deep}, SatMAE~\cite{cong2022satmae}, Scale-MAE~\cite{reed2023scale}, DOFA~\cite{xiong2024neural}}\\
        (ii) & \multicolumn{6}{l}{DuPLo~\cite{interdonato2019duplo}, Transformer~\cite{russwurm2020self}, LTAE~\cite{garnot2020lightweight}, TempCNN~\cite{pelletier2019temporal}} \\
        (iii) & \multicolumn{6}{l}{UTAE~\cite{garnot2021panoptic}, TSViT (seg. head)~\cite{tarasiou2023vits}}\\
        (iv) & \multicolumn{6}{l}{\{PSE~\cite{garnot2020satellite}, SatMAE~\cite{cong2022satmae}, Scale-MAE~\cite{reed2023scale}, DOFA~\cite{xiong2024neural}\}+LTAE~\cite{garnot2020lightweight}, TSViT (cls. head)~\cite{tarasiou2023vits}}\\
        \bottomrule
    \end{tabular}
    }
    \label{tab:baselines}
\end{table}

\end{document}